\documentclass[10pt,twocolumn,letterpaper]{article}
\usepackage{multirow}
\usepackage{graphicx}
\usepackage{subcaption}
\usepackage{bm}
\usepackage[table]{xcolor}
\usepackage{soul}
\usepackage{makecell}
\usepackage{cvpr}              

\definecolor{cvprblue}{rgb}{0.21,0.49,0.74}
\usepackage[pagebackref,breaklinks,colorlinks,citecolor=cvprblue]{hyperref}

\def\paperID{6010} 
\def\confName{CVPR}
\def\confYear{2024}

\title{Harnessing the Power of MLLMs for Transferable Text-to-Image Person ReID}

\author{
    {Wentao Tan$^{1,3}$~~~~~
    Changxing Ding$^{1,2}$\thanks{Corresponding author}~~~~~
    Jiayu Jiang$^1$~~~~~
    Fei Wang$^{1,3}$~~~~~
    Yibing Zhan$^3$~~~~~
    Dapeng Tao$^{4,5}$} 
    \\
    $^1$South China University of Technology \quad 
    $^2$Pazhou Lab, Guangzhou \quad
    $^3$JD Explore Academy, Beijing \quad
    \\
    $^4$Yunnan University \quad
    $^5$Yunnan United Vision Technology Co., Ltd., Kunming
    \\
    {\tt\small \{ftwentaotan,202320111494,ft\_feiw\}@mail.scut.edu.cn, chxding@scut.edu.cn}
    \\
    {\tt\small zhanyibing@jd.com, dapeng.tao@gmail.com} 
    \\
    {\tt\small \textcolor{blue}{\url{https://github.com/WentaoTan/MLLM4Text-ReID}}}
}

\begin{document}
\begin{sloppypar}
\maketitle
\begin{abstract}
Text-to-image person re-identification (ReID) retrieves pedestrian images according to textual descriptions. Manually annotating textual descriptions is time-consuming, restricting the scale of existing datasets and therefore the generalization ability of ReID models. As a result, we study the transferable text-to-image ReID problem, where we train a model on our proposed large-scale database and directly deploy it to various datasets for evaluation. We obtain substantial training data via Multi-modal Large Language Models (MLLMs). Moreover, we identify and address two key challenges in utilizing the obtained textual descriptions. First, an MLLM tends to generate descriptions with similar structures, causing the model to overfit specific sentence patterns. Thus, we propose a novel method that uses MLLMs to caption images according to various templates. These templates are obtained using a multi-turn dialogue with a Large Language Model (LLM). Therefore, we can build a large-scale dataset with diverse textual descriptions. Second, an MLLM may produce incorrect descriptions. Hence, we introduce a novel method that automatically identifies words in a description that do not correspond with the image. This method is based on the similarity between one text and all patch token embeddings in the image. Then, we mask these words with a larger probability in the subsequent training epoch, alleviating the impact of noisy textual descriptions. The experimental results demonstrate that our methods significantly boost the direct transfer text-to-image ReID performance. Benefiting from the pre-trained model weights, we also achieve state-of-the-art performance in the traditional evaluation settings. 
\end{abstract}    
\vspace{-1em}
\section{Introduction}\label{sec:intro}

Text-to-image person re-identification (ReID) \cite{wang2020vitaa,ding2021semantically,shao2022learning,farooq2022axm,yan2023learning, jiang2023cross,shao2023unified,wang2022uncertainty,tan2023style,ding2020multi,7917252} is a task that retrieves pedestrian images according to textual descriptions. It is a powerful tool when probe images of the target person are unavailable and only textual descriptions exist. It has various potential applications, including video surveillance \cite{bukhari2023language}, social media analysis \cite{li2017person}, and crowd management \cite{galiyawala2021person}. However, it remains challenging mainly because annotating textual descriptions for pedestrian images is time-consuming \cite{yang2023towards}. Consequently, existing datasets \cite{li2017person,ding2021semantically,zhu2021dssl} for text-to-image person ReID are usually small, resulting in insufficient deep model training.

\begin{figure}[t]
\vspace{-1em}
\centerline{\includegraphics[width=1.0\linewidth]{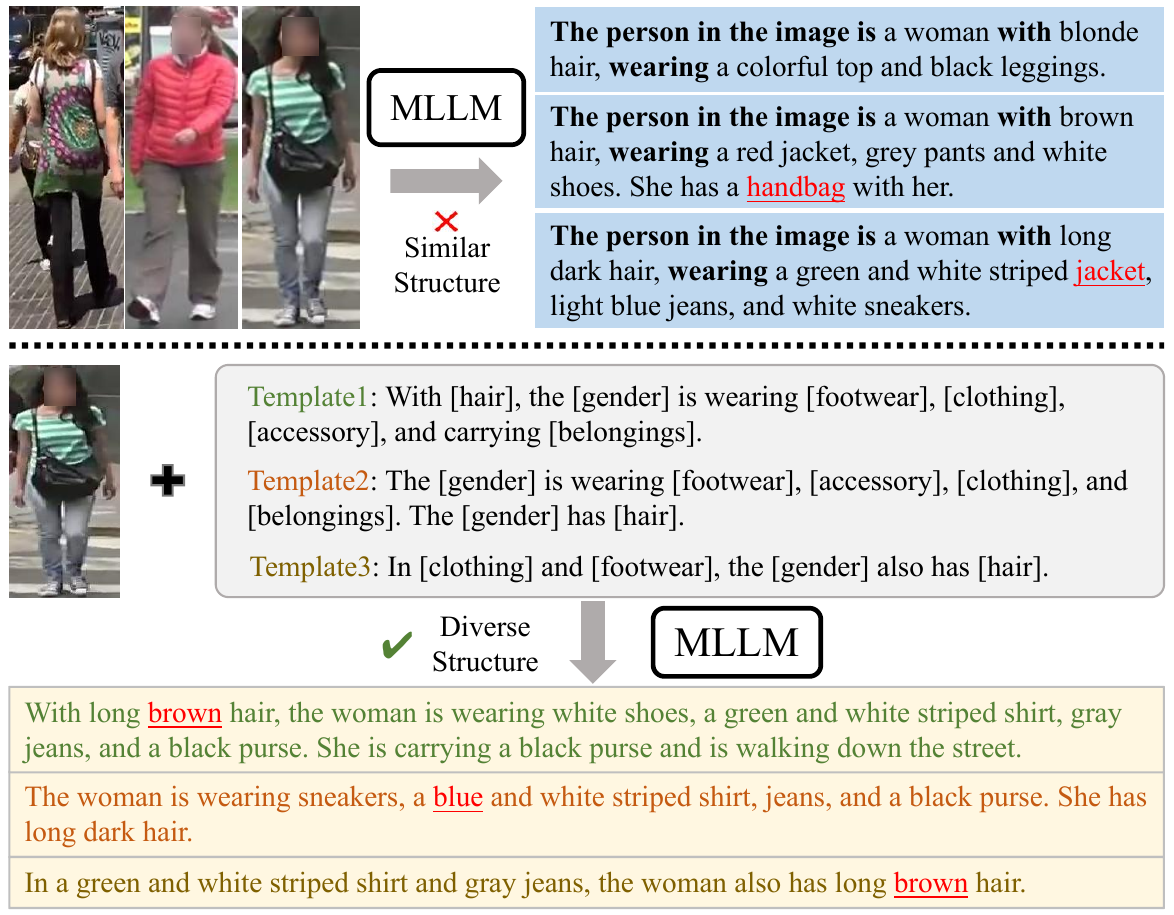}}
  \caption{Illustration of textual descriptions generated by an MLLM (\textit{i.e.}, Qwen \cite{bai2023qwen}). (Top) The description patterns are similar for different images. (Bottom) Our proposed Template-based Diversity Enhancement (TDE) method significantly enhances the description pattern diversity. It is worth noting that some errors are present in the generated descriptions shown in this figure.}
  \label{fig:similar_caption}
  \vspace{-1em}
\end{figure}

Previous studies on text-to-image ReID usually assumed that training and testing data are drawn from the same domain. They proposed novel model architectures \cite{ding2021semantically,shao2022learning,radford2021learning,bai2023rasa,xie2020learning,xie2021viewpoint}, loss functions \cite{zhang2018deep,zheng2020dual,xie2023towards}, and pre-training strategies \cite{shao2023unified,yang2023towards} to improve model performance for each database. However, researchers have recently discovered that the cross-dataset generalization ability of their approaches is significantly low \cite{shao2022learning}, limiting real-world applications. Since annotating textual descriptions is time-consuming, collecting training data for each target domain is infeasible. Therefore, training a model that can be directly deployed to various target domains is necessary.

Accordingly, we study the transferable text-to-image ReID problem. The term ``transferable" is derived from the seminal work CLIP~\cite{radford2021learning}, which refers to a large-scale pre-trained model's capacity that directly applies its knowledge to other domains or tasks without fine-tuning on labeled data. 
Due to the rapid advancements in multi-modal large language models (MLLMs) \cite{yin2023survey,bai2023qwen,chen2023shikra,li2022blip}, we utilize them to generate textual descriptions automatically and employ them to replace traditional manual annotations. Specifically, we utilize the large-scale LUPerson dataset \cite{fu2021unsupervised} as the image source and generate textual descriptions using MLLMs. The obtained image-text pairs are utilized to train a model directly evaluated in existing text-to-image ReID databases. However, to improve the model's transfer ability, two essential challenges must be addressed: (1) guiding MLLMs to generate diverse textual descriptions for a single image and (2) reducing the impact of the noise in the synthesized textual descriptions. 

First, MLLMs tend to generate descriptions with similar sentence structures, as shown in Fig. \ref{fig:similar_caption}. This causes the text-to-image ReID model to overfit specific sentence patterns, reducing the model's ability to generalize to various human description styles encountered in real-world applications.  To address this issue, we propose a Template-based Diversity Enhancement (TDE) method that instructs MLLMs to conduct image captioning according to given description templates. Obtaining these templates with minimal effort involves performing multi-turn dialogues with ChatGPT \cite{ChatGPT} and prompting it to generate diverse templates. Then, we randomly integrate one of these templates into the MLLM's captioning instruction, resulting in vivid descriptions with varied sentence structures. This approach significantly enhances textual description diversity.

Second, although MLLMs are highly effective, the generated descriptions still contain errors. This implies that certain words in a textual description may not match the paired image. Thus, we propose a novel Noise-aware Masking (NAM) method to address this problem. Specifically, we compute the similarities between each text token and all image tokens in the paired image for a specific textual description. The similarity scores between the unmatched word and image tokens are usually low. Hence, we identify potentially incorrect words and mask them with a large probability in the next training epoch before they are fed into a text encoder. Furthermore, NAM and Masked Language Modeling (MLM) are similar but have two key differences: (1) MLM masks all tokens with equal probability, while NAM masks them based on their noise level. (2) MLM applies cross-entropy loss to predict the masked tokens, whereas NAM focuses on masking words without predicting potentially noisy words. In the experimentation section, we demonstrate NAM's ability to effectively alleviate the impact of noisy textual descriptions.

To the best of our knowledge, this is the first study focusing on the transferable text-to-image ReID problem by harnessing the power of MLLMs. We innovatively generate diverse textual descriptions and minimize the impact of the noise contained in these descriptions. The experimental results show that our method performs excellently on three popular benchmarks in both direct transfer and traditional evaluation settings.

\section{Related Works}
\label{sec:related_works}

\textbf{Text-to-Image Re-Identification.}
Existing approaches for this task improve model performance from three perspectives: model backbone \cite{bai2023rasa,jiang2023cross}, feature alignment strategies \cite{jiang2023cross,shao2022learning,zhang2018deep}, and pre-training \cite{shao2023unified,yang2023towards}. 

The first method category improves the model backbone. Early approaches adopted the VGG model \cite{li2017person,chen2018improving} and LSTM \cite{memory2010long,zhang2018deep,yan2023image} as image and text encoders, respectively. These encoders gradually evolve into ResNet-50 \cite{he2016deep,ding2021semantically,wang2020vitaa,farooq2022axm} and BERT \cite{sarafianos2019adversarial,devlin2018bert,zhu2021dssl,shu2022see,li2022learning} models. Moreover, the CLIP \cite{radford2021learning,han2021text} and ALBEF-based encoders \cite{li2021align,bai2023rasa,yang2023towards} have recently become popular. Notably, the CLIP model contains jointly pre-trained image and text encoders. Thus, its cross-modal alignment capabilities are advantageous and have proven more effective than the individually pre-trained encoders \cite{shao2023unified}. Moreover, the ALBEF model \cite{li2021align} performs interaction between visual and textual features, which improves the feature representation capacity but brings in significant computational cost.

The second category of methods enhances feature alignment strategies. Previous methods aligned an image's holistic features with its textual description \cite{wang2019language,shu2022see,wang2016learning,zhang2018deep,aggarwal2020text,wu2023refined,xie2021object}. Subsequent approaches \cite{lee2018stacked,niu2020improving,jing2020pose,wang2022caibc,chen2022tipcb,gao2021contextual,wang2022look,tang2022learning,li2023knowledge} focused on aligning the image-text pair's local features to suit the fine-grained retrieval nature of text-to-image ReID. These approaches can be divided into explicit and implicit alignment methods. Explicit methods \cite{wang2020vitaa,ding2021semantically} extract the visual- and textual-part features and then compute the alignment loss between them. Implicit methods can also align local features \cite{farooq2022axm,shao2022learning,yang2023towards}. For example, Jiang et al. \cite{jiang2023cross} applied MLM to text tokens and then predicted the masked tokens using image token features. This indirectly realizes local feature alignment between the image patch and noun phrase representations. 

Since existing databases are small, two recent studies explored pre-training for text-to-image ReID. Shao \textit{et al.} \cite{shao2023unified} utilized the CLIP model to predict the attributes of a pedestrian image. Then, they inserted these attributes into manually defined description templates. As a result, they obtained a large number of pre-training data. Similarly, Yang \textit{et al.} \cite{yang2023towards} utilized the text descriptions from the CUHK-PEDES \cite{li2017person} and ICFG-PEDES \cite{ding2021semantically} datasets to synthesize images using a diffusion model \cite{rombach2022high}. Then, they used the BLIP model \cite{li2022blip} to caption these images and obtain a large-scale pre-training dataset. However, these two studies targeted at pre-training and did not investigate the direct transfer setting where no target domain data is available for fine-tuning. Moreover, they overlooked the noise or diversity issues generated in the obtained textual descriptions.

The above methods achieve excellent in-domain performance; however, their cross-dataset performance is usually significantly low \cite{shao2022learning}. This paper explores the transferable text-to-image ReID task with minimal manual operations. Also, we address the challenges in textual descriptions generated by MLLMs.

\textbf{Multi-modal Large Language Models.}
Multi-modal Large Language Models (MLLMs) \cite{yin2023survey,touvron2023llama,touvron2023llama2,zhu2023minigpt,liu2023visual} are built on Large Language Models (LLMs) \cite{zhao2023survey,yang2023baichuan,brown2020language,zheng2023judging,chowdhery2022palm} and incorporate textual and non-textual information as input \cite{bai2023qwen,chen2023shikra,han2023imagebind}. This paper only considers MLLMs that use both texts and images as input signals. The input text (\textit{i.e.}, the ``instruction" or ``prompt") describes the tasks assigned to MLLMs to understand the image's content.
Regarding MLLM architecture, most studies \cite{li2023blip2,wang2022git,liu2023visual} first map the image patch and text token embeddings into a shared feature space and then perform decoding using a LLM. Some methods \cite{alayrac2022flamingo} improve the interaction and alignment strategies between the image and text tokens during decoding, facilitating more stable training \cite{li2023multimodal}.

In this paper, we utilize MLLMs to eliminate the need to manually annotate textual descriptions. We also explore strategies to address the diversity and noise issues in the obtained textual descriptions, facilitating the development of a transferable text-to-image ReID model.

\section{Methods}

\begin{figure*}[t]
\vspace{-1em}
\centering
\includegraphics[width=1.0\linewidth]{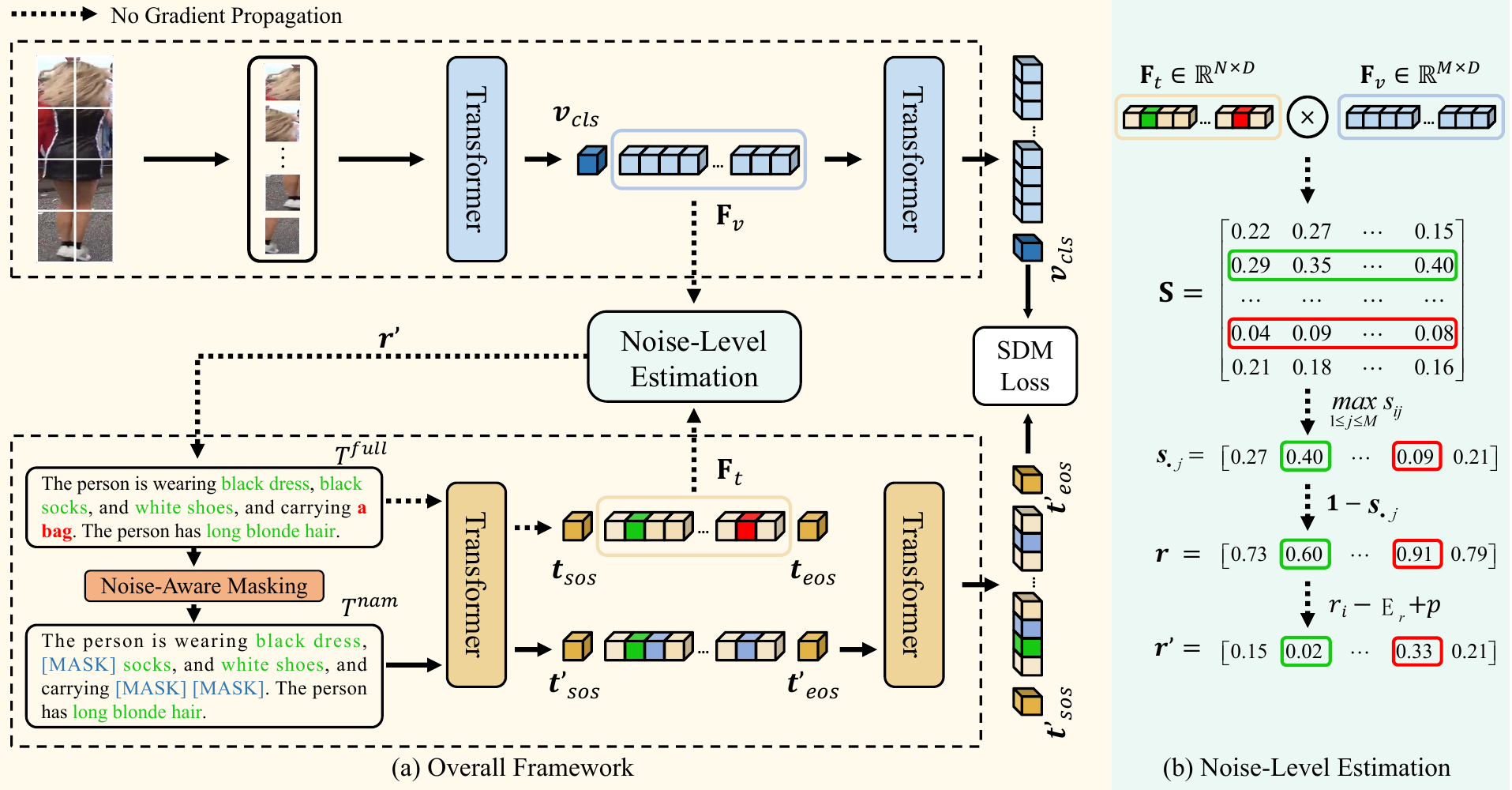}
\caption{Overview of our framework. We adopt the CLIP-ViT/B-16 model as the backbone. Our framework uses one pedestrian image, the original textual description $T^{full}$, and a masked textual description $T^{nam}$ as input during training. $T^{nam}$ is obtained by applying NAM to $T^{full}$. To perform NAM, we first compute the similarity matrix $\mathbf{S}$ between the text 
tokens $\mathbf{F_\textit{t}}$ of $T^{full}$ and the image tokens $\mathbf{F_\textit{v}}$ according to their embeddings at the $l$-th layer of the encoders. Then, we estimate the probability of each text token's noisiness according to the similarity between its embedding and the image token embeddings. The similarity distribution matching (SDM) loss is computed between the global visual feature $\bm{v}_{cls}$ of the pedestrian image and the global textual feature $\bm{t'}_{eos}$ of $T^{nam}$. The model's optimization quality is enhanced by masking noisy words in $T^{full}$. (Best viewed in color.)}

\label{fig:framewrok}
\end{figure*}

The overview of our solution to the transferable text-to-image ReID problem is illustrated in Fig. \ref{fig:framewrok}.
Section \ref{sec:mllm} addresses diversity issues associated with textual descriptions generated by MLLMs. Section \ref{sec:nam} discusses the reduction of noise impact in the descriptions. And section \ref{sec:optimization} outlines the loss function utilized for model optimization.

\subsection{Generating Diverse Descriptions} \label{sec:mllm}

Manually annotating textual descriptions for pedestrian images is time-consuming and hardly scalable. Fortunately, MLLMs have advanced rapidly and provide effective image captioning. Therefore, we decide to utilize MLLMs to create large-scale text annotations for training a model with excellent transfer capacity.

\textbf{Instruction Design.}
We adopt the LUPerson database \cite{fu2021unsupervised} as the image source because it holds a significant amount of images that were captured in diverse environments. 
A technical aspect of using MLLMs lies in designing an effective instruction, which usually depends on user experience. We solve this problem using a multi-turn dialogue with ChatGPT \cite{ChatGPT}, and this process is detailed in the supplementary material. The resulting instruction is as follows:

\emph{
  ``Write a description about the overall appearance of the person in the image, including the attributes: clothing, shoes, hairstyle, gender and belongings. If any attribute is not visible, you can ignore it. Do not imagine any contents that are not in the image."
}

This is considered a \emph{static instruction} as it is fixed for all images. In this paper, the textual descriptions generated using the \emph{static instruction} are denoted as \emph{static texts} or ${T^{s}}$.

\textbf{Diversity Enhancement.} An MLLM generates textual descriptions with similar sentence patterns for different images using the \emph{static instruction}, as illustrated in Fig. \ref{fig:similar_caption}. This causes the text-to-image ReID model to overfit these sentence patterns, limiting its generalization to real-world descriptions. We attempt to improve the \emph{static instruction}, but the obtained sentence patterns remained limited.
Although using more MLLMs can bring in multiple sentence patterns, these patterns are still far from diverse.

Again, we resort to ChatGPT to solve this problem. Specifically, we propose a Template-based Diversity Enhancement (TDE) method. 
First, we generate two descriptions for each of a set of images using two MLLMs \cite{bai2023qwen,chen2023shikra} according to the \emph{static instruction}. Then, we feed these descriptions to ChatGPT to capture their sentence patterns (\textit{i.e.}, description templates). With the guidance of these templates, we instruct ChatGPT to create more templates. Finally, it produces 46 templates after multi-turn dialogues, which are detailed in the supplementary material. We randomly select one of the templates and insert it into the \emph{static instruction}, obtaining a \emph{dynamic instruction} as follows:
 
\emph{``Generate a description about the overall appearance of the person, including clothing, shoes, hairstyle, gender, and belongings, in a style similar to the template: `\{template\}'. If some requirements in the template are not visible, you can ignore them. Do not imagine any contents that are not in the image."} 

The \emph{`\{template\}'} is replaceable. Furthermore, the textual descriptions generated according to the \emph{dynamic instruction} are referred to as \emph{dynamic texts} (${T^{d}}$).
As illustrated in Fig. \ref{fig:similar_caption}, MLLMs can follow the sentence patterns specified in the templates, significantly enhancing the diversity of the obtained textual descriptions.

\textbf{Dataset Description.} We utilize the publicly available Qwen \cite{bai2023qwen} and Shikra \cite{chen2023shikra} models in this paper. By harnessing the power of the two MLLMs, we obtain the large-scale LUPerson-MLLM dataset. This dataset comprises 1.0 million images, and each image has four captions, $T^s_{qwen}$, $T^s_{shikra}$, $T^d_{qwen}$, and $T^d_{shikra}$. The first and the last two captions are generated according to the static and dynamic instructions, respectively. We reserve the $T^s$ for each image as we observe that its description is usually complementary to that of $T^d$. In the following section, we will train the model with LUPerson-MLLM. For simplicity, we refer all the above MLLM-generated descriptions as $T^{full}$.

\subsection{Noise-Aware Masking} \label{sec:nam}
Although MLLMs are powerful, they cannot describe images very precisely. As depicted in Fig. \ref{fig:similar_caption} and Fig. \ref{fig:framewrok}, a few words do not match the described image in the obtained textual descriptions. Existing methods \cite{he2022synthetic,li2022blip} usually discard the noisy descriptions, losing the other valuable information contained in the matched words. Accordingly, we propose a novel noise-aware masking (NAM) method that identifies noisy text tokens and fully uses the matched text tokens for model training. 


\textbf{Image Encoder.} 
An image is divided into $M$ non-overlapped patches. These image tokens are concatenated with the [CLS] token and are fed into the image encoder. Then, the [CLS] token embedding at the last image encoder layer is used as the global image feature, denoted as $\bm{v}_{cls} \in \mathbb{R}^d$. The feature dimension is represented by $d$.


\textbf{Text Encoder.} We tokenize each textual description $T^{full}$ into a sequence of  $N$ tokens. The $N$ of each sentence varies according to its length. The token sequence is bracketed with [SOS] and [EOS] to represent the start and the end of the sequence. Meanwhile, we examine each text token's noise level in $T^{full}$, which is computed  and stored in the previous training epoch. These values are used to perform NAM on $T^{full}$ to obtain $T^{nam}$. After that, $T^{full}$ and $T^{nam}$ are fed into the text encoder independently. At the final text encoder layer, the global feature $\bm{t'}_{eos}$ of $T^{nam}$ is utilized to calculate loss. $T^{full}$ is only used for NAM, which means it is not used for loss computation.

\textbf{Noise-Aware Masking.} We utilize the image and text encoders' token embeddings in the $l$-th layers for the noise-level estimation of $T^{full}$. These embeddings are denoted as 
$\mathbf{F_\textit{v}} = [\bm{v}^l_1, ..., \bm{v}^l_M]$ and $\mathbf{F_\textit{t}} = [\bm{t}^l_1, ..., \bm{t}^l_N]$, respectively, where $\bm{v}^l_j \in \mathbb{R}^d$ and $\bm{t}^l_j \in \mathbb{R}^d$.

Furthermore, we calculate the token-wise similarity between a single text-image pair as follows:
\begin{align} 
 \mathbf{S} = \mathbf{F_\textit{t}}^T \mathbf{F_\textit{v}} ,
\end{align}
where $\mathbf{S} \in \mathbb{R}^{N \times M}$ is a similarity matrix and $s_{ij}$ represents the cosine similarity between the $i$-th text token embedding and the $j$-th image token embedding. If one text token does not match the image, the similarity scores between this token's embedding and those of all the image tokens will be consistently be low. Therefore, the noise level of the $i$-th text token in $T^{full}$ can be estimated via:
\begin{align} \label{eq:noise}
r_i = 1 - (\max_{1 \leq j \leq M} \bm{s}_{ij}).
\end{align}
By applying Eq.\eqref{eq:noise} to each row of $\mathbf{S}$, we obtain a vector $ \bm{r} = [r_1, ..., r_N]$ that records the noise-level of all text tokens.

Moreover, NAM applies the masking operation to all the text tokens in $T^{full}$ with different probabilities, which can be determined based on the noise-level values recorded in $\bm{r}$. However, in the initial training stage, the  values of elements in $\bm{r}$ may be high. This results in excessive masking of important tokens and hinders learning. To resolve this issue, we modify the expectation value of all $\bm{r}$ elements into a constant number as described below:
\begin{align}
    \mathbb{E}_r = \frac{1}{N} \sum_{i=1}^{N} r_i, 
\end{align} 
\begin{align}
  \bm{r'} = [r_1 -\mathbb{E}_r+p ,..., r_N-\mathbb{E}_r+p],
\end{align} 
where $p$ is the average masking ratio. We utilize the $\bm{r'}$ values as the final probability that a text token might be masked. We include the pseudo code and visualization of NAM in the supplementary materials.



\textbf{Discussion.} 
Computing $\bm{r'}$ and then applying NAM to obtain $T^{nam}$ in each iteration requires two forward passes. This additional time cost cannot be overlooked in large-scale training. In contrast, our strategy computes $\bm{r'}$ for the next training epoch, which requires only one forward pass for each iteration. Furthermore, we initialize the $\bm{r'}$ values with the constant $p$ in the first training epoch. 

\subsection{Optimization} \label{sec:optimization}
Following \cite{jiang2023cross}, we adopt the similarity distribution matching (SDM) loss to optimize our model. Given a mini-batch of $B$ matched image-text pairs $\{(\bm{v}^i_{cls}, \bm{t}^{'i}_{eos})\}^B_i$, we first establish the matching relationship between each image and text (\textit{i.e.}, $\{(\bm{v}_{cls}^i, \bm{t}^{'j}_{eos}), y_{i,j}\} (1 \leq i,j \leq B) $), where $y_{i,j} = 1$ and $y_{i,j} = 0$ denote a positive and a negative image-text pair, respectively. 
Then, we calculate the ground truth matching distribution $\mathbf{q_\textit{i}}$ for the $i$-th image, where its $j$-th element is $q_{i,j} = y_{i,j} /\sum_{b=1}^{B}y_{i,b}$. Finally,
we align the predicted probability distribution $\mathbf{p_\textit{i}}$ with $\mathbf{q_\textit{i}}$ as follows:

\begin{equation} \label{eq:sdm_i2t}
  \mathcal{L}_{i2t} = \frac{1}{B} \sum_{i=1}^{B} KL(\mathbf{p_\textit{i}}\| \mathbf{q_\textit{i}}) = \frac{1}{B} \sum_{i=1}^{B}\sum_{j=1}^{B}p_{i,j}\log(\frac{p_{i,j}}{q_{i,j} + \epsilon}), 
\end{equation}
where $\epsilon$ is a small number to avoid numerical problems and 
\begin{equation} \label{eq:logit_i2t}
  p_{i,j} = \frac{\exp(sim(\bm{v}_{cls}^i, \bm{t}^{'j}_{eos})/\tau )}{\sum_{b=1}^{B} \exp(sim(\bm{v}_{cls}^i, \bm{t}^{'b}_{eos})/\tau)}.
\end{equation}
$sim(\mathbf{u, v}) = \mathbf{u^\top v}/\|\mathbf{u}\| \|\mathbf{v}\| $ denotes the cosine similarity between $\mathbf{u}$ and $\mathbf{v}$, $\tau$ is a temperature coefficient. 

The SDM loss from text to image $\mathcal{L}_{t2i}$ can be computed by exchanging the position of $\bm{v}_{cls}$ and $\bm{t}^{'}_{eos}$ in Eq.~\eqref{eq:sdm_i2t} and Eq.~\eqref{eq:logit_i2t}. Finally, the complete SDM loss is computed as follows:
\begin{equation} \label{eq:sdm_loss}
  \mathcal{L}_{sdm} = \mathcal{L}_{i2t} + \mathcal{L}_{t2i}. 
\end{equation}
It is worth noting that since we randomly sample images from the large-scale LUPerson database, we assume that each image in a sampled batch has a unique identity.

\section{Experiments}

\begin{table*}[t]
\caption{Ablation study on each key component in the direct transfer setting. `CLIP' refers to directly using the original CLIP encoders provided in \cite{radford2021learning}.}
\vspace{-1em}
\label{tab:ablation}

\centering
\resizebox{0.95\linewidth}{!}{
\begin{tabular}{c||cc|cc|c|ccc|ccc|ccc}
\toprule[1pt]
\multirow{2}{*}{Method} & \multirow{2}{*}{$T^s_{qwen}$} & \multirow{2}{*}{$T^s_{shikra}$} & \multirow{2}{*}{$T^d_{qwen}$} & \multirow{2}{*}{$T^d_{shikra}$} & \multirow{2}{*}{NAM} & \multicolumn{3}{c|}{CUHK-PEDES} & \multicolumn{3}{c|}{ICFG-PEDES} & \multicolumn{3}{c}{RSTPReID} \\ \cline{7-15} 
 &  &  &  &  &  & R1 & R5 & mAP & R1 & R5 & mAP & R1 & R5 & mAP \\ \hline
CLIP &  &  &  &  &  & 12.65 & 27.16 & 11.15 & 6.67 & 17.91 & 2.51 & 13.45 & 33.85 & 10.31 \\ \hline
\multirow{3}{*}{Static Text} & $\checkmark$ &  &  &  &  & 37.65 & 57.86 & 33.40 & 23.78 & 42.77 & 11.18 & 36.30 & 60.60 & 26.25 \\
 &  & $\checkmark$ &  &  &  & 39.70 & 62.60 & 36.09 & 19.02 & 35.63 & 9.67 & 36.90 & 62.65 & 28.33 \\
 & $\checkmark$& $\checkmark$&  &  &  & 46.00 & 66.82 & 41.27 & 26.74 & 44.22 & 13.23 & 41.10 & 66.95 & 30.21 \\ \hline
\multirow{3}{*}{Dynamic Text} &  &  & $\checkmark$&  &  & 40.72 & 62.36 & 37.21 & 24.16 & 41.24 & 11.32 & 38.65 & 64.70 & 28.81 \\
 &  &  &  & $\checkmark$ &  & 43.63 & 65.46 & 39.08 & 22.07 & 39.57 & 11.35 & 38.80 & 63.45 & 28.60 \\
 &  &  & $\checkmark$ & $\checkmark$ &  & 48.86 & 69.41 & 44.09 & 28.43 & 46.37 & 14.23 & 44.25 & 66.15 & 32.99 \\ \hline
TDE & $\checkmark$& $\checkmark$& $\checkmark$& $\checkmark$&  & 50.32 & 71.36 & 45.74 & 29.12 & 47.96 & 15.13 & 45.70 & 70.75 & 33.23 \\ \hline
NAM & $\checkmark$& $\checkmark$& $\checkmark$& $\checkmark$& $\checkmark$& 52.64 & 71.62 & 46.48 & 32.61 & 50.79 & 16.48 & 47.75 & 70.75 & 34.73 \\ \hline
\end{tabular}
}
\end{table*}

\subsection{Datasets and Settings} \label{sec:dataset}

\textbf{CUHK-PEDES.} CUHK-PEDES \cite{li2017person} is a pioneer dataset in the text-to-image ReID field. Each image in this dataset has two textual descriptions. The training set comprises data on 11,003 identities, including 34,054 images and 68,108 textual descriptions. In contrast, the testing set contains 3,074 images and 6,156 textual descriptions from 1,000 identities.

\textbf{ICFG-PEDES.}
ICFG-PEDES \cite{ding2021semantically} contains of 54,522 images from 4,102 identities. Each image has one textual description. The training set consists of 34,674 image-text pairs corresponding to 3,102 identities, while the testing set comprises 19,848 image-text pairs from the remaining 1,000 identities.

\textbf{RSTPReid.}
RSTPReid \cite{zhu2021dssl} includes 20,505 images captured by 15 cameras from 4,101 identities. 
Each identity has five images captured with different cameras and each image has two textual descriptions. According to the official data division, the training set incorporates data from 3,701 identities, while both the validation and testing sets include data from 200 identities, respectively.

\textbf{LUPerson.} LUPerson \cite{fu2021unsupervised} contains 4,180,243 pedestrian images sampled from 46,260 online videos, covering a variety of scenes and  view points. The images are from over 200K pedestrians. 

\textbf{Evaluation Metrics.}
Like existing works \cite{jiang2023cross,bai2023rasa,shao2023unified,yang2023towards}, we adopt the popular Rank-\textit{k} accuracy (\textit{k}=1,5,10) and mean Average Precision (mAP) as the evaluation metrics for the three databases. Moreover, we consider the following two evaluation settings.

\textbf{Direct Transfer Setting.} For this setting, the model is only trained on the LUPerson-MLLM dataset, and the above three benchmarks are tested immediately. This setting directly evaluates the quality of our dataset and the effectiveness of the proposed methods (\textit{i.e.}, TDE and NAM).

\textbf{Fine-tuning Setting.} In this setting, we first pre-train our model on the LUPerson-MLLM dataset and then fine-tune it on each of the three benchmarks respectively. 


\subsection{Implementation Details}
Similar to previous studies \cite{jiang2023cross,chen2023towards}, we adopt CLIP-VIT-B/16 \cite{radford2021learning} as the image encoder and a 12-layer transformer as our text encoder. The input image resolution is resized to 384 $\times$ 128 pixels. Additionally, we apply random horizontal flipping, random cropping, and random erasing as data augmentation for the input images. Each textual description is first tokenized, with a maximum length of 77 tokens (including the [SOS] and [EOS] tokens). The hyper-parameter $p$ is set to 0.15 and the temperature coefficient $\tau$ in Eq.~\eqref{eq:logit_i2t} is set to 0.02. The model is trained using the Adam optimizer with a learning rate of 1e-5 and cosine learning rate decay strategy. We train each model on 8 TITAN-V GPUs, with 64 images per GPU. The training process lasts for 30 epochs. The versions of the mentioned LLM/MLLMs are ChatGPT-3.5-Turbo, Qwen-VL-Chat-7B, and Shikra-7B.

\subsection{Ablation Study}
We randomly sample 0.1 million images from our LUPerson-MLLM database to accelerate the ablation study on the direct transfer evaluation setting. Then, we increase the amount of training images to 1.0 million to enhance the transfer ability of our text-to-image ReID models.

\textbf{Effectiveness of TDE.}
The experiments in Table \ref{tab:ablation} show that \emph{dynamic instruction} is better than \emph{static instruction}. For example, the model using only $T^d_{qwen}$ outperforms that the one using $T^s_{qwen}$ by about 3\% in Rank-1 performance on the CUHK-PEDES database. On the same database and evaluation metric, the model that uses only $T^d_{shikra}$ outperforms the one using $T^s_{shikra}$ by about 4\%. These experimental results indicate that enhancing sentence pattern diversity improves the transfer ability of ReID models. Therefore, we use the four descriptions for each image in the subsequent experiments. It is worth noting that none of the above experiments employ NAM. Instead, they mask every text token with an equal probability of $p$.

\textbf{Effectiveness of NAM.}
MLLM-generated textual descriptions often contain noise, which is harmful for model training. Replacing the equal masking strategy with our NAM method improves our model's Rank-1 performance by 2.32\%, 3.49\%, and 2.05\% on the three databases, respectively. These improvements are even higher than the benefits of combining dynamic and static texts (\textit{i.e.}, 1.46\%, 0.69\%, and 1.45\%). These experimental results demonstrate that NAM identifies the noisy words in the text and effectively reduces their impact. NAM allows the model to accurately align visual and textual features, thereby enhancing the direct transfer text-image ReID performance.

\begin{figure}[t]
  \centerline{\includegraphics[width=1.0\linewidth]{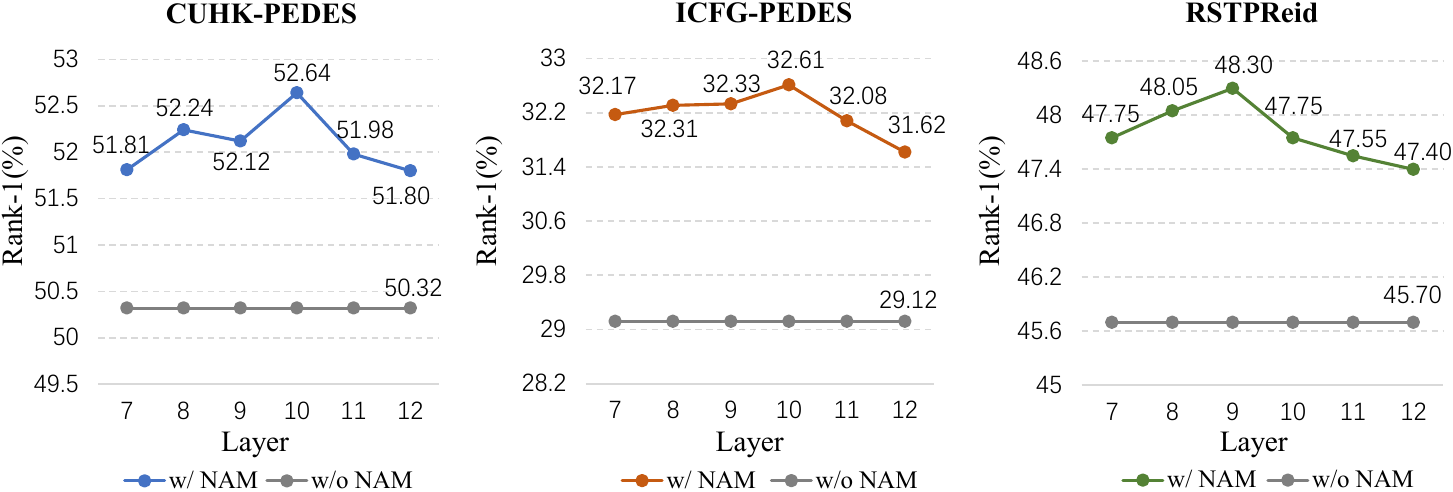}}
  \caption{Results of different layers for NAM to compute $S$. The encoders contain 12 layers in total. Best viewed with zoom-in.}
  \label{fig:layer}
\end{figure}

\textbf{The Layer where NAM Computes $\mathbf{S}$.}
$\mathbf{S}$ contains pairwise similarity scores between features in $ \mathbf{F_\textit{v}}$ and $ \mathbf{F_\textit{t}}$. This experiment investigates the optimal layer for obtaining $ \mathbf{F_\textit{v}}$ and $ \mathbf{F_\textit{t}}$. The results are plotted in Fig. \ref{fig:layer}. We observe that the model's performance consistently improves regardless of the layer used to provide $ \mathbf{F_\textit{v}}$ and $ \mathbf{F_\textit{t}}$. We also notice that the adopted encoders' $10$-th layer yields the best overall performance. Compared to the last encoder layer, the $10$-th layer may offer more fine-grained information, facilitating more accurate similarity computation between token pairs.

\begin{figure}[t]
  \centerline{\includegraphics[width=1.0\linewidth]{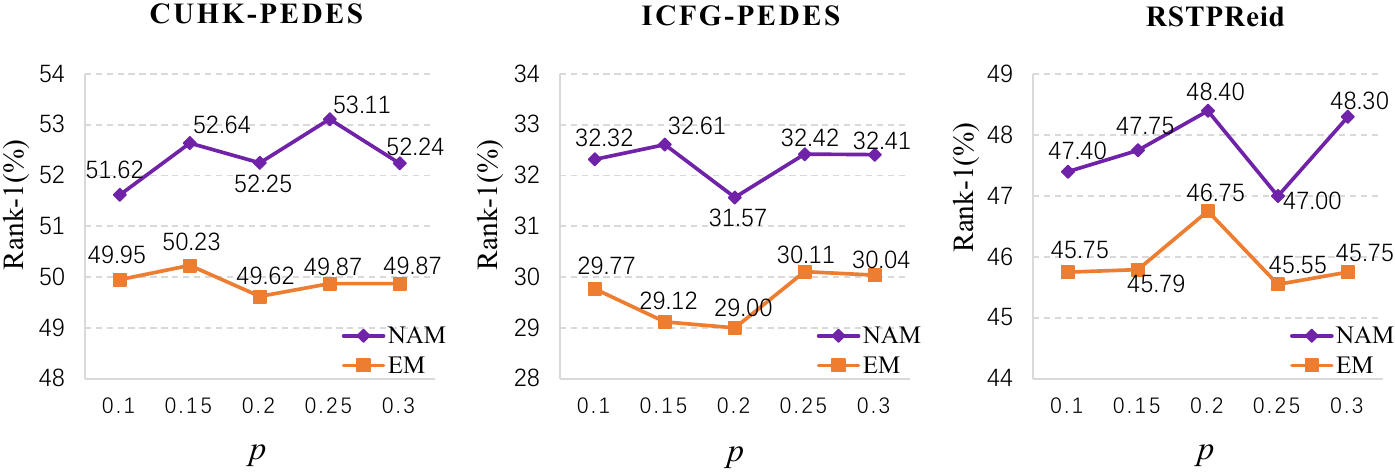}}
  \caption{Results of different overall masking ratios $p$ for NAM. `EM' represents masking all text tokens with the same probability $p$. Best viewed with zoom-in. }
  \label{fig:nam_p}
\end{figure}

\textbf{The Overall Masking Ratio for NAM.}
Our NAM method masks different text tokens with unequal probabilities, but it maintains an overall probability of $p$. In this experiment, we explore the optimal $p$ value. To demonstrate NAM's advantages, we also include the results of the masking tokens with equal probabilities (referred to as “EM"). As shown in Table \ref{fig:nam_p}, NAM consistently outperforms EM with various $p$ values. The optimal value of $p$ is about 0.15.


\textbf{Combination of NAM and MLM.} 
MLM requires the model to predict the masked text tokens. It has proven effective and is widely applied in NLP models. Recent text-to-image ReID studies \cite{jiang2023cross} confirm that MLM loss is beneficial when the textual descriptions are manually annotated. However, our NAM doesn't predict the masked tokens as the textual descriptions generated by MLLMs may be noisy. Table \ref{tab:nam_vs_mlm} shows that applying MLM loss to NAM is harmful, indicating the MLLM description noise is a crucial issue.

\begin{figure}[t]
  \centerline{\includegraphics[width=0.8\linewidth]{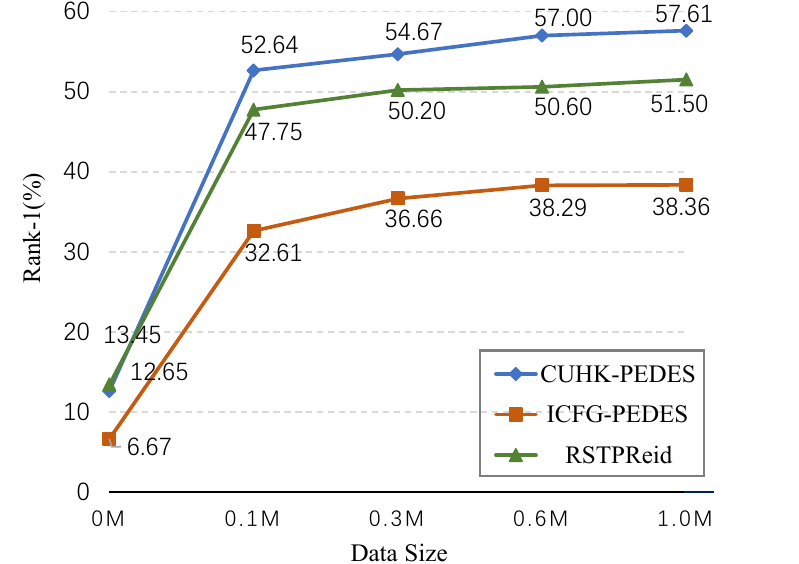}}
  \caption{Training data size's impact on our methods' direct transfer ReID performance. `0 M' refers to directly using the original CLIP encoders.}
  \label{fig:scale}
\end{figure}

\textbf{The Data Size Impact.}
The dataset size is essential to training. More pre-trained data improves the performance. 
We investigate the effect of training data size on the direct transfer ReID performance and summarize the results in Fig. \ref{fig:scale}. It is evident that the model's direct transfer performance steadily improves as the data amount increases. Finally, compared with the model using only 0.1 million training images, the Rank-1 performance of the model using 1.0 million training images is significantly promoted by 5.75\% on the challenging ICFG-PEDES database, indicating that our approach can scale to large-scale database.

\begin{table}[tp]
\caption{Results of the combination of NAM and the MLM loss.}
\vspace{-1em}
\label{tab:nam_vs_mlm}
\centering
\resizebox{0.45\textwidth}{!}{%
\begin{tabular}{c||cc|cc|cc}
\toprule[1pt]
\multirow{2}{*}{Method} & \multicolumn{2}{c|}{CUHK-PEDES} & \multicolumn{2}{c|}{ICFG-PEDES} & \multicolumn{2}{c}{RSTPReid} \\ \cline{2-7} 
 & R1 & mAP & R1 & mAP & R1 & mAP \\ \hline \hline
EM & 50.32 & 45.74 & 29.12 & 15.13 & 45.70 & 33.23  \\ 
NAM & 52.64 & 46.48 & 32.61 & 16.48 & 47.75 & 34.73  \\
NAM w/ MLM loss & 48.79 & 43.86 & 27.36 & 14.16 & 44.45 & 33.07 \\ \hline
\end{tabular}}
\end{table}

\begin{table}[tp]
\caption{Comparisons with existing pre-training datasets in the direct transfer setting.}
\vspace{-1em}
\label{tab:sota-zeroshot}
\centering
\resizebox{0.45\textwidth}{!}{%
\begin{tabular}{c||cc|cc|cc}
\toprule[1pt]
\multirow{2}{*}{Pretrain Dataset} & \multicolumn{2}{c|}{CUHK-PEDES} & \multicolumn{2}{c|}{ICFG-PEDES} & \multicolumn{2}{c}{RSTPReid} \\ \cline{2-7} 
 & R1 & mAP & R1 & mAP & R1 & mAP \\ \hline \hline
None & 12.65 & 11.15 & 6.67 & 2.51 & 13.45 & 10.31 \\ \hline
MALS \cite{yang2023towards} (1.5 M) & 19.36 & 18.62 & 7.93 & 3.52 & 22.85 & 17.11 \\ \hline
LUPerson-T \cite{shao2023unified} (0.95 M) & 21.88 & 19.96 & 11.46 & 4.56 & 22.40 & 17.08 \\ \hline
Ours (0.1 M) & 52.64  & 46.48  &32.61  & 16.53 &47.75  &34.73  \\
Ours (1.0 M) & 57.61  & 51.44 &38.36  &20.43  & 51.50 & 37.34 \\ \hline
\end{tabular}}
\end{table}

\begin{table}[tp]
\caption{Comparisons with existing pre-training datasets in the fine-tuning setting.}
\vspace{-1em}
\label{tab:sota-finetune}

\centering
\resizebox{0.45\textwidth}{!}{%
\begin{tabular}{c||c|clclcl}
\toprule[1pt]
\multirow{3}{*}{Init Parameters} & \multirow{3}{*}{Source} & \multicolumn{6}{c}{Target} \\ \cline{3-8} 
 &  & \multicolumn{2}{c|}{CUHK-PEDES} & \multicolumn{2}{c|}{ICFG-PEDES} & \multicolumn{2}{c}{RSTPReid} \\ \cline{3-8} 
 &  & R1 & \multicolumn{1}{l|}{mAP} & R1 & \multicolumn{1}{l|}{mAP} & R1 & mAP \\ \hline \hline

\multirow{3}{*}{CLIP \cite{radford2021learning}} & CUHK-PEDES & \cellcolor{gray!30}{73.48} & \multicolumn{1}{l|}{\cellcolor{gray!30}{66.21}} & 43.04 & \multicolumn{1}{l|}{22.45} & 52.55 & 39.97 \\
 & ICFG-PEDES & 33.90 & \multicolumn{1}{l|}{31.65} & \cellcolor{gray!30}{63.83} & \multicolumn{1}{l|}{\cellcolor{gray!30}{38.37}} & 47.45 & 36.83 \\
 & PSTPReid & 35.25 & \multicolumn{1}{l|}{32.35} & 33.58 & \multicolumn{1}{l|}{19.58} & \cellcolor{gray!30}{60.40} & \cellcolor{gray!30}{47.70} \\ \hline
\multirow{3}{*}{\makecell[c]{MALS \cite{yang2023towards} \\ (1.5 M)}} & CUHK-PEDES & \cellcolor{gray!30}{74.05} & \multicolumn{1}{l|}{\cellcolor{gray!30}{66.57}} & 44.53 & \multicolumn{1}{l|}{22.66} & 53.55 & 39.17 \\
 & ICFG-PEDES & 40.38 & \multicolumn{1}{l|}{36.83} & \cellcolor{gray!30}{64.37} & \multicolumn{1}{l|}{\cellcolor{gray!30}{38.85}} & 49.00 & 38.20 \\
 & PSTPReid & 38.40 & \multicolumn{1}{l|}{34.47} & 34.11 & \multicolumn{1}{l|}{20.82} & \cellcolor{gray!30}{61.90} & \cellcolor{gray!30}{48.08} \\ \hline
\multirow{3}{*}{ \makecell[c]{LuPerson-T \cite{shao2023unified} \\ (0.95 M)}} & CUHK-PEDES & \cellcolor{gray!30}{74.37} & \multicolumn{1}{l|}{\cellcolor{gray!30}{66.60}} & 44.30 & \multicolumn{1}{l|}{22.67} & 53.75 & 38.98 \\
 & ICFG-PEDES & 35.07 & \multicolumn{1}{l|}{32.47} & \cellcolor{gray!30}{64.50} & \multicolumn{1}{l|}{\cellcolor{gray!30}{38.22}} & 48.05 & 38.21 \\
 & PSTPReid & 38.29 & \multicolumn{1}{l|}{34.43} & 35.81 & \multicolumn{1}{l|}{21.62} & \cellcolor{gray!30}{62.20} & \cellcolor{gray!30}{48.33} \\ \hline
\multirow{3}{*}{Ours (0.1 M)} & CUHK-PEDES & \cellcolor{gray!30}{\textbf{74.64}} & \multicolumn{1}{l|}{\cellcolor{gray!30}{\textbf{67.44}}} & \textbf{46.19} & \multicolumn{1}{l|}{\textbf{24.08}} & \textbf{56.15} & \textbf{40.84} \\
 & ICFG-PEDES & \textbf{56.70} & \multicolumn{1}{l|}{\textbf{51.23}} & \cellcolor{gray!30}{\textbf{65.30}} & \multicolumn{1}{l|}{\cellcolor{gray!30}{\textbf{39.90}}} &\textbf{52.60} & \textbf{39.76} \\
 & PSTPReid & \textbf{56.69} & \multicolumn{1}{l|}{\textbf{51.40}} & \textbf{42.70} & \multicolumn{1}{l|}{\textbf{25.69}} & \cellcolor{gray!30}{\textbf{64.05}} & \cellcolor{gray!30}{\textbf{49.27}} \\ \hline
\multirow{3}{*}{Ours (1.0 M)} & CUHK-PEDES & \cellcolor{gray!30}{\textbf{76.82}} & \multicolumn{1}{l|}{\cellcolor{gray!30}{\textbf{69.55}}} & \textbf{49.38} & \multicolumn{1}{l|}{\textbf{26.92}} & \textbf{59.60} & \textbf{44.70} \\
 & ICFG-PEDES & \textbf{61.20} & \multicolumn{1}{l|}{\textbf{55.60}} & \cellcolor{gray!30}{\textbf{67.05}} & \multicolumn{1}{l|}{\cellcolor{gray!30}{\textbf{41.51}}} &\textbf{54.80} & \textbf{42.56} \\
 & PSTPReid & \textbf{62.99} & \multicolumn{1}{l|}{\textbf{57.20}} & \textbf{48.44} & \multicolumn{1}{l|}{\textbf{30.03}} & \cellcolor{gray!30}{\textbf{68.50}} & \cellcolor{gray!30}{\textbf{53.02}} \\ \hline
 
\end{tabular}
}
\end{table}

\begin{table*}[tp]
\caption{Comparisons with state-of-the-art methods in the traditional evaluation settings.}
\vspace{-1em}
\label{tab:sota-traditional}

\centering
\resizebox{0.975\linewidth}{!}{
\begin{tabular}{l||cc|cccc|cccc|cccc}
\toprule[1pt]
\multirow{2}{*}{Method} & \multirow{2}{*}{Image Enc.} & \multirow{2}{*}{Text Enc.} & \multicolumn{4}{c|}{CUHK-PEDES} & \multicolumn{4}{c|}{ICFG-PEDES}                    & \multicolumn{4}{c}{RSTPReid}  \\ \cline{4-15} 
                        &                             &                            & R1     & R5     & R10   & mAP   & R1                         & R5    & R10   & mAP   & R1    & R5    & R10   & mAP   \\ \hline \hline

CMPM/C \cite{zhang2018deep}                  & RN50                        & LSTM                       & 49.37  & -      & 79.27 & -     & 43.51                      & 65.44 & 74.26 & -     & -     & -     & -     & -     \\
ViTAA \cite{wang2020vitaa}                   & RN50                        & LSTM                       & 55.97  & 75.84  & 83.52 & -     & 50.98                      & 68.79 & 75.78 & -     & -     & -     & -     & -     \\
DSSL \cite{zheng2020dual}                    & RN50                        & BERT                       & 59.98  & 80.41  & 87.56 & -     & -                          & -     & -     & -     & 32.43 & 55.08 & 63.19 & -     \\ 
SSAN \cite{ding2021semantically}                    & RN50                        & LSTM                       & 61.37  & 80.15  & 86.73 & -     & 54.23 & 72.63 & 79.53 & -     & 43.50 & 67.80 & 77.15 & -     \\ 
LapsCore \cite{wu2021lapscore}               & RN50                        & BERT                       & 63.40  & -      & 87.80 & -     & -                          & -     & -     & -     & -     & -     & -     & -     \\
LBUL \cite{wang2022look}                    & RN50                        & BERT                       & 64.04  & 82.66  & 87.22 & -     & -                          & -     & -     & -     & 45.55 & 68.2  & 77.85 & -     \\
SAF \cite{li2022learning}                     & ViT-Base                    & BERT                       & 64.13  & 82.62  & 88.4  & -     & -                          & -     & -     & -     & -     & -     & -     & -     \\
TIPCB \cite{chen2022tipcb}                   & RN50                        & BERT                       & 64.26  & 83.19  & 89.1  & -     & 54.96                      & 74.72 & 81.89 & -     & -     & -     & -     & -     \\
CAIBC \cite{wang2022caibc}                   & RN50                        & BERT                       & 64.43  & 82.87  & 88.37 & -     & -                          & -     & -     & -     & 47.35 & 69.55 & 79.00 & -     \\
AXM-Net \cite{farooq2022axm}                 & RN50                        & BERT                       & 64.44  & 80.52  & 86.77 & 58.70 & -                          & -     & -     & -     & -     & -     & -     & -     \\
LGUR \cite{shao2022learning}                    & DeiT-Small                  & BERT                       & 65.25  & 83.12  & 89.00 & -     & 59.02                      & 75.32 & 81.56 & -     & 47.95 & 71.85 & 80.25 & -     \\
IVT \cite{shu2022see}                     & ViT-Base                    & BERT                       & 65.69  & 85.93  & 91.15 & -     & 56.04                      & 73.60 & 80.22 & -     & 46.70 & 70.00 & 78.80 & -     \\
LCR²S \cite{yan2023learning}                   & RN50                        & TextCNN+BERT               & 67.36  & 84.19  & 89.62 & 59.20 & 57.93                      & 76.08 & 82.40 & 38.21 & 54.95 & 76.65 & 84.70 & 40.92 \\
UniPT \cite{shao2023unified}                   & ViT-Base                    & BERT                       & 68.50  & 84.67  & 90.38 & -     & 60.09                      & 76.19 & 82.46 & -     & 51.85 & 74.85 & 82.85 & -     \\
\midrule
\multicolumn{6}{l}{\textit{with CLIP \cite{radford2021learning} backbone:}} \\
\midrule
Han et al. \cite{han2021text} & CLIP-RN101 & CLIP-Xformer & 64.08 & 81.73 & 88.19 & 60.08 & - & - & - & - & - & - & - & - \\
IRRA \cite{jiang2023cross}                    & CLIP-ViT                    & CLIP-Xformer               & 73.38  & 89.93  & 93.71 & 66.10 & 63.46                      & 80.25 & 85.82 & 38.06 & 60.20 & 81.30 & 88.20 & 47.17 \\
MALS \cite{yang2023towards} + IRRA        & CLIP-ViT                    & CLIP-Xformer               &74.05        & 89.48       & 93.64      & 66.57      & 64.37                           & 80.75      & 86.12      &38.85       &61.90       &80.60       &89.30       & 48.08 \\
LUPerson-T \cite{shao2023unified} + IRRA           & CLIP-ViT                    & CLIP-Xformer               & 74.37       &89.51        & 93.97      & 66.60      & 64.50                           &80.24       & 85.74      &38.22      & 62.20      &83.30       & 89.75      &48.33  \\

Ours (1.0 M) + IRRA          & CLIP-ViT                    & CLIP-Xformer               &     \textbf{76.82}   &   \textbf{91.16}     &  \textbf{94.46}     &  \textbf{69.55}     &          \textbf{67.05}                  &  \textbf{82.16}     & \textbf{87.33}      & \textbf{41.51}      & \textbf{68.50}      & \textbf{87.15}      &\textbf{92.10}       & \textbf{53.02}       \\ 
\midrule
\multicolumn{6}{l}{\textit{with ALBEF \cite{li2021align} backbone:}} \\
\midrule

RaSa \cite{bai2023rasa}                    & CLIP-ViT                      & BERT-base                     & 76.51  & 90.29  & 94.25 & \textbf{69.38} & 65.28  & 80.40  & 85.12 & 41.29 & 66.90 & 86.50 & 91.35 & 52.31 \\
APTM \cite{yang2023towards}                    & Swin-B                   & BERT-base                  & 76.53  & 90.04  & 94.15 & 66.91 & 68.51                      & 82.99 & 87.56 & 41.22 & 67.50 & 85.70 & 91.45 & 52.56 \\
Ours (1.0 M) + APTM  & Swin-B & BERT-base & \textbf{78.13}  & \textbf{91.19} & \textbf{94.50} & 68.75 &\textbf{69.37}  &\textbf{83.55}  &\textbf{88.18}  &\textbf{42.42}  &\textbf{69.95}  &\textbf{87.35}  &\textbf{92.30}  &\textbf{54.17}  \\ \hline



\end{tabular}}
\end{table*}

\subsection{Comparisons with State-of-the-Art Methods}

\textbf{Comparisons with Other Pre-training Datasets.}
MALS \cite{yang2023towards} and LUPerson-T \cite{shao2023unified} are two pre-training datasets in the field of text-to-image ReID. MALS \cite{yang2023towards} contains 1.5 M images, with textual descriptions obtained using the BLIP model \cite{li2022blip}. However, it does not address the diversity and noise issues in the obtained descriptions. LUPerson-T \cite{shao2023unified} contains 0.95 M images that were also sampled from the LUPerson database \cite{shao2023unified}. It utilizes the CLIP model to predict pedestrian attributes and inserts them into manually defined templates as textual descriptions. We utilize the three databases to train the CLIP-ViT/B-16 model, incorporating the SDM loss. Finally, we evaluate the model's performance in both direct transfer and fine-tuning settings.

Comparisons on the direct transfer setting are summarized in Table \ref{tab:sota-zeroshot}. It is shown that the model trained on the LUPerson-MLLM dataset achieves significantly better performance, even when we only sample 0.1 M images. This is because TDE enables diverse description generation. Moreover, NAM efficiently alleviates the impact of noise in textual descriptions. 
Combining both techniques results in a model that exhibits exceptional transfer abilities.
In comparison, neither \cite{yang2023towards} nor \cite{shao2023unified} consider the noise problem in their obtained textual descriptions.

Table \ref{tab:sota-finetune} displays the model comparisons in the fine-tuning setting. In this experiment, we adopt the IRRA method \cite{jiang2023cross} in the fine-tuning stage and initialize its parameters with each of the above three pre-trained models, respectively. The fine-tuned models are evaluated on both in-domain and cross-domain text-to-image ReID scenarios. According to the results in Table \ref{tab:sota-finetune}, two conclusions can be derived. First, compared with the CLIP model \cite{radford2021learning}, pre-training using the three pre-training datasets exhibits performance promotion for in-domain and cross-domain tasks. Second, pre-training using LUPerson-MLLM exhibits the most remarkable performance promotion. For example, in the ICFG-PEDES $\rightarrow$ CUHK-PEDES setting, 
LUPerson-MLLM outperforms the other two models by 20.82\% and 26.13\% in Rank-1 accuracy, respectively. These experimental results further validate the effectiveness of our  methods.

\textbf{Comparisons in the Traditional Evaluation Settings.}
Comparisons with state-of-the-art approaches are summarized in Table \ref{tab:sota-traditional}. We observe that our method achieves the best performance.
With our pre-trained model parameters, the Rank-1 accuracy and mAP of IRRA are improved by 8.30\% and 5.85\% on the RSTPReid database, respectively. Besides, pre-training with our LUPerson-MLLM dataset is more effective than with the MALS and LUPerson-T datasets. This is because we effectively resolve the diversity and noise issues in the MLLM descriptions, facilitating more robust and discriminative feature learning.

\section{Conclusion and Limitations}
This paper explores the challenging transferable text-to-image ReID problem by harnessing the image captioning capability of MLLMs. We acknowledge diversity and noise as critical issues in utilizing the obtained textual descriptions. To address these two problems, we introduce the Template-based Diversity Enhancement (TDE) method to encourage diverse description generation and construct a large-scale dataset named LUPerson-MLLM. In addition, we proposed the NAM method to mitigate the impact of noisy textual descriptions. Extensive experiments demonstrate that TDE and NAM significantly improve the model's transfer power. However, these methods have limitations: the effectiveness of TDE is limited by the number of sentence templates; NAM may occasionally fail to mask noisy tokens. In the future, we aim to explore more powerful methods to address diversity and noise issues in MLLM-generated descriptions.

\textbf{Broader Impacts.} TDE addresses fixed sentence patterns generated by MLLMs, inspiring effective instruction design to harness MLLMs' capabilities. Meanwhile, NAM tackles text noise generated by MLLMs, facilitating wider MLLM adoption for practical real-world problems.

\textbf{Acknowledgement.} This work was partially supported by the Major Science and Technology Innovation 2030 ``New Generation Artificial Intelligence” key project (No. 2021ZD0111700), the National Natural Science Foundation of China under Grants 62076101 and 62172354, the Guangdong Basic and Applied Basic Research Foundation under Grant 2023A1515010007, the Guangdong Provincial Key Laboratory of Human Digital Twin under Grant 2022B1212010004, and the Yunnan Provincial Major Science and Technology Special Plan Projects under Grant 202202AD080003. We also gratefully acknowledge the support and resources provided by the Yunnan Key Laboratory of Media Convergence, the CAAI Huawei MindSpore Open Fund and the TCL Young Scholars Program.

\clearpage
{
    \small
    \bibliographystyle{ieeenat_fullname}
    \bibliography{main}
}

\end{sloppypar}
\end{document}